\documentclass[runningheads]{llncs}
\usepackage[T1]{fontenc}
\usepackage{graphicx}
\usepackage{booktabs}
\usepackage{amssymb}
\usepackage{amsmath} 
\usepackage{multicol}
\usepackage{multirow}
\begin{document}
\title{Can Multimodal Large Language Model Think Analogically?}
\author{Diandian Guo\inst{1,2} \orcidID{0009-0002-8468-3285}\and
Cong Cao\thanks{Corresponding author.} \inst{1} \and
Fangfang Yuan\inst{1} \and Dakui Wang\inst{1} \and Wei Ma\inst{1} \and Yanbing Liu\inst{1,2} \and Jianhui Fu\inst{3}
}
\authorrunning{D. Guo et al.}
\institute{Institute of Information Engineering, Chinese Academy of Sciences  \and
School of Cyber Security, University of Chinese Academy of Sciences \\
\email{\{guodiandian, caocong, yuanfangfang, wangdakui, mawei, liuyanbing\}@iie.ac.cn}\\ 
\and  Shandong Institutes of Industrial Technology \\
\email{fujianhui2020@qq.com}
}
\maketitle              % typeset the header of the contribution
\begin{abstract}
Analogical reasoning, particularly in multimodal contexts, is the foundation of human perception and creativity. Multimodal Large Language Model (MLLM) has recently sparked considerable discussion due to its emergent capabilities. In this paper, we delve into the multimodal analogical reasoning capability of MLLM. Specifically, we explore two facets: \textit{MLLM as an explainer} and \textit{MLLM as a predictor}. In \textit{MLLM as an explainer}, we primarily focus on whether MLLM can deeply comprehend multimodal analogical reasoning problems. We propose a unified prompt template and a method for harnessing the comprehension capabilities of MLLM to augment existing models. In \textit{MLLM as a predictor}, we aim to determine whether MLLM can directly solve multimodal analogical reasoning problems.  The experiments show that our approach outperforms existing methods on popular datasets, providing preliminary evidence for the analogical reasoning capability of MLLM.

\end{abstract}
\section{Introduction}
Analogical reasoning - the ability to perceive and use relational similarity between two situations or events - serves as a fundamental pillar in human cognition and creativity \cite{gentner2012analogical}. It constitutes a critical mechanism for discerning complex relations \cite{meguro2020effects}, facilitating abstract concept comprehension \cite{glynn1989analogical}, and fostering innovative problem-solving capabilities \cite{novick2005problem}. From scientific discoveries \cite{dunbar2012scientific} to everyday decision-making \cite{cattell1971abilities}, the capacity for analogical reasoning plays an indispensable role in the cognitive toolkit of human intellect.

Researchers in deep learning have consistently endeavored to investigate methodologies for endowing models with human-like capabilities \cite{hu2023thought}. Recently, there has been considerable exploratory work on whether it is possible to capture analogical reasoning abilities in deep learning systems \cite{mitchell2021abstraction}.  Ethayarajh \textit{et al.} \cite{ethayarajh2018towards} devote to word analogy recognition, which can be effectively solved by word embeddings. Some studies have further evaluated the analogical thinking ability of pre-trained language models \cite{brown2020language,ushio2021bert}. The latest research \cite{webb2023emergent,yasunaga2023large} provides preliminary evidence that Large Language Model (LLM) possesses analogical reasoning abilities. Meanwhile, many attempts \cite{zhang2019raven,hu2021stratified,malkinski2022deep} in visual analogical reasoning primarily focus on integrating relational, structural, and analogical reasoning to enhance model intelligence. 

\begin{figure}[tbp]
\begin{center}
\includegraphics[width=0.6\linewidth]{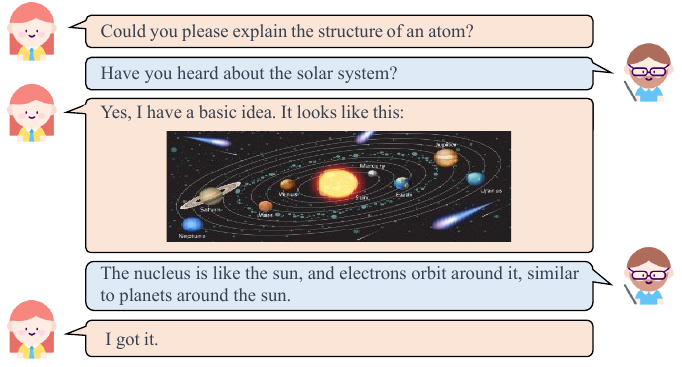} 
\caption{Humans often establish initial cognitive understanding through multimodal analogical reasoning when dealing with unfamiliar problems. One analogical reasoning example is $Sun : Solar\;system :: Nucleus : Atom. $
}
\label{fig:motivation}
\end{center}
\end{figure}
In practical scenarios, humans typically employ experiential knowledge (such as visual information) to engage in analogical reasoning when confronted with unfamiliar problems, thereby establishing a preliminary understanding of those problems. As illustrated in Figure \ref{fig:motivation}, this type of reasoning is often multimodal. However, existing research on analogical reasoning predominantly focuses on single modality, with limited attention dedicated to studying multimodal contexts. Multimodal Large Language Model (MLLM) has recently emerged as a prominent research focus, leveraging powerful LLMs as the core mechanism to execute multimodal tasks \cite{yin2023survey}. MLLMs have learned extensive relational patternsduring self-supervised learning, which can identify and utilize these patterns without explicit training in analogical reasoning. Therefore, we aim to explore whether MLLM possesses the capability for multimodal analogical reasoning, offering a new perspective for evaluating MLLM.

In this paper, we explore the application of MLLM in multimodal analogical reasoning task from two perspectives: \textit{MLLM as an explainer} and \textit{MLLM as a predictor}. In \textit{MLLM as an explainer}, our primary focus lies on MLLM's capacity to comprehend and describe multimodal analogical reasoning problems. We aim to enhance the performance of existing methods in multimodal analogical reasoning task by providing elaborate explanations generated by MLLM. Specifically, we unify the prompt template used in existing Multimodal Pre-trained Transformer (MPT) methods, employ MLLM to explain multimodal analogical reasoning problems, and then incorporate the explanations into the corresponding slots within the templates. On the other hand, in \textit{MLLM as a predictor}, we mainly investigate whether MLLM itself can solve multimodal analogical reasoning problems, aiming to explore its intuitive reasoning capabilities. To achieve this, we structure multimodal analogical reasoning task in a natural language format tailored to MLLM and design a two-step fine-tuning framework. The first step fine-tuning aims to enable MLLM to learn background triplet knowledge, while the second step fine-tuning aims to teach MLLM the format of multimodal analogical reasoning task. To summarize, our main contributions are the following:
\begin{itemize}
    \item To our best knowledge, we are the first to explore the multimodal analogical reasoning capabilities of MLLM from two perspectives: \textit{MLLM as an explainer} and \textit{MLLM as a predictor}.

    \item Experimental results demonstrate that our proposed approaches achieve state-of-the-art performance, which preliminarily proves that MLLM has multimodal analogical reasoning capability.
\end{itemize}

\section{Related Work}
\paragraph{\textbf{Multimodal Knowledge Graph Embedding (MKGE)}} \;
Although MKGE methods can not directly complete multimodal analogical reasoning task, they can accomplish this task by decomposing it into a pipeline form of \textit{relation prediction $\rightarrow$ template filling $\rightarrow$ entity prediction}. Existing methods primarily focus on encoding image features into knowledge graph embeddings. For instance, IKRL \cite{Xie_Liu_Luan_Sun_2017} extends TransE \cite{bordes2013translating} by modeling visual representations from both entity and structural information. TranAE \cite{Wang_Li_Li_Zeng_2019} and MoSE \cite{zhao2022mose} enable different modalities to be represented in the same embedding space. RSME \cite{Wang_Wang_Yang_Zhang_Chen_Qi_2021} focuses on noisy images that do not correspond to target entities and select high-quality images. These methods emphasize structured information but suffer from limited scalability and require model structure redesign for different tasks.

\paragraph{\textbf{Multimodal Pre-trained Model (MPM)}} \; MPMs have recently demonstrated great superiority in many multimodal tasks. We divide MPMs into two categories: MPT and MLLM. MPT can accomplish multimodal analogical reasoning task by constructing prompts and predicting the $\mathrm{[MASK]}$ token, including: \textbf{1. single-stream models}, such as VisualBERT \cite{li2019visualbert} and ViLT \cite{kim2021vilt}, where image and textual embeddings are combined into a sequence and passed through a transformer to obtain contextual representations. \textbf{2. dual-stream models}, like ViLBERT \cite{lu2019vilbert}, which interact through transformer layers with cross-modal or shared attention, separating visual and language processing into two streams. \textbf{3. mixed-stream models}, including FLAVA \cite{singh2022flava} and MKGformer \cite{chen2022hybrid}, which leverage a unified framework to conduct various multimodal tasks. Recently, with the popularity of LLMs, research on MLLMs has also been increasing. For example, there are models like VisualGLM \cite{du2022glm}, which is based on ChatGLM \cite{zeng2022glm} and LLaVA \cite{liu2023llava}, which is based on Llama \cite{touvron2023llama}. However, there is still a lack of exploration into the analogical reasoning capabilities of MLLM.

\begin{figure}[ht]
\begin{center}
\includegraphics[width=0.9\linewidth]{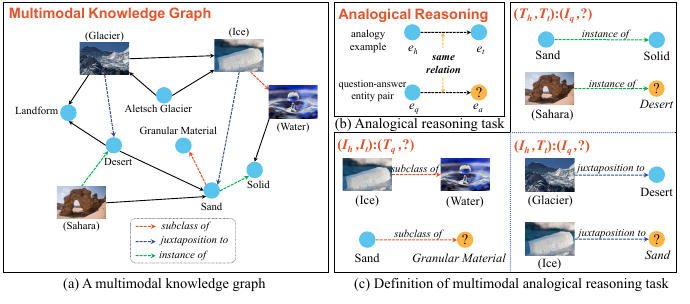} 
\caption{Overview of the multimodal analogical reasoning task. We provide a multimodal knowledge graph in Figure \ref{fig:def}(a). Figure \ref{fig:def}(b) shows a general analogical reasoning task. Figure \ref{fig:def}(c) shows three subtasks of multimodal analogical reasoning task. Please note that the relations indicated by dashed arrows ($\dashrightarrow$) and the text in parentheses beneath the images are \textbf{for annotation purposes only} and are \textbf{not included} in the input.}
\label{fig:def}
\end{center}
\end{figure}
\section{Methodology}
In this section, we first introduce the multimodal analogical reasoning task and propose two viable frameworks, namely \textit{MLLM as an explainer} and \textit{MLLM as a predictor}.

\subsection{Task Definition}
\label{def}
As illustrated in Figure \ref{fig:def}(b), the conventional analogical reasoning task can be formalized as $(e_h,e_t) : (e_q,?)$. Analogical reasoning, given an analogy example $(e_h, e_t)$ and a question-answer entity pair $(e_q, ?)$ with $e_h, e_t, e_q \in \mathcal{E}$, aims to predict the missing entity $e_a \in \mathcal{E}$. It is important to note that the relations between $(e_h,e_t)$ and $(e_q,e_a)$ are identical but not currently available. 

Multimodal analogical reasoning is first proposed by \cite{zhang2023multimodal}, and is based on the background multimodal knowledge graph $\mathcal{G}=(\mathcal{E}, \mathcal{R}, \mathcal{I}, \mathcal{T}, \mathcal{S})$. Here, $\mathcal{E}$ and $\mathcal{R}$ are the entity set and relation set, respectively, $\mathcal{I}$ and $\mathcal{T}$ represent images and textual descriptions of entities, and $\mathcal{S}$ is the triplet set. Multimodal analogical reasoning tasks follow the form of the conventional analogical reasoning task. Based on the different modalities of $e_h$, $e_t$ and $e_q$, it can be divided into three subtasks: %Given an analogy example $(e_h,e_t)$ and a question-answer entity pair $(e_q,?)$, where $e_h,e_t,e_q \in \mathcal{E}$, the goal is to predict the missing entity $e_a \in \mathcal{E}$. Note that the relations of $(e_h,e_t)$ and $(e_q,e_a)$ are identical but unavailable.

%Multimodal analogical reasoning task is first proposed by  \cite{zhang2023multimodal}, which is based on the background multimodal knowledge graph $\mathcal{G}=(\mathcal{E}, \mathcal{R}, \mathcal{I}, \mathcal{T}, \mathcal{S})$, where $\mathcal{E}$ and $\mathcal{R}$ are entity set and relation set, $\mathcal{I}$ and $\mathcal{T}$ represent images and textual descriptions of entities, $\mathcal{S}$ is triplet set. Multimodal analogical reasoning task also follows the form of $(e_h,e_t) : (e_q,?)$. As shown in Figure \ref{fig:def}(c), this task can be further divided into three subtasks according to different modalities of $e_h$,$e_t$ and $e_q$:

\begin{itemize}
    \item $(I_h,I_t):(T_q,?)$ \; The modalities of $(e_h,e_t)$ are visual, while the modality of the question entity $e_q$ is textual.
    \item $(T_h,T_t):(I_q,?)$ \; The modalities of $(e_h,e_t)$ are textual, while the modality of the question entity $e_q$ is visual.
    \item $(I_h,T_t):(I_q,?)$ \; We set the modalities of $e_h$ and $e_q$ to visual and the modality of $e_t$ to textual.
    
\end{itemize}

\begin{figure}[htbp]
\begin{center}
\includegraphics[width=0.8\linewidth]{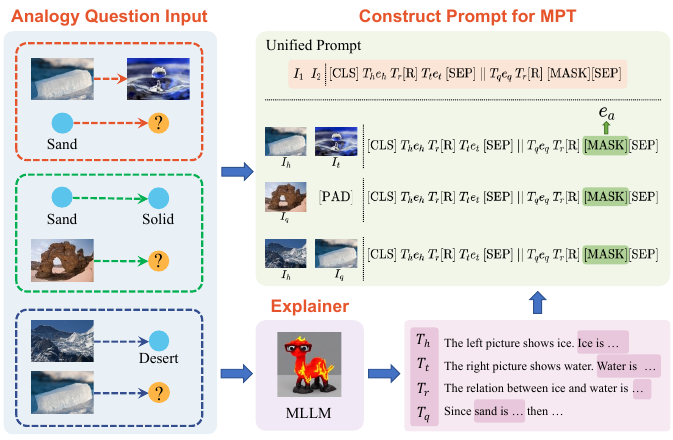} 
\caption{MLLM as an explainer.}
\label{fig:m1}
\end{center}
\end{figure}

\subsection{MLLM as an Explainer}
For \textit{MLLM as an explainer}, our approach is to explore the powerful comprehension abilities of MLLM to enhance the performance of the MPT methods on multimodal analogical reasoning task. The overall framework of \textit{MLLM as an explainer} is shown in Figure \ref{fig:m1}.

\subsubsection{Unified Prompt Template}
Existing MPT models use different prompt templates for three distinct subtasks when performing multimodal analogical reasoning task, which is unnecessary and redundant. Therefore, we propose a unified prompt template as follows:

\begin{equation}
\begin{split}
     \mathcal{T} =  &I_1 \; I_2 \; \mathrm{[CLS]} \; T_h e_h \;T_r \mathrm{[R]}\; T_t e_t\; \mathrm{[SEP]} \; || \\ &T_q e_q \;T_r \mathrm{[R]\;[MASK]\;[SEP]} 
\end{split}
\end{equation}
where $I$ represents the given images of the entity; $T_h,T_t$ and $T_q$ are the textual descriptions of the entities $e_h,e_t$ and $e_q$; $T_r$ is the textual description of the implicit relation; $||$ is the concatenate operation. As the relations are not explicitly given in this task, we designate $\mathrm{[R]}$ as a special token to explicitly indicate the relation. MPT models are trained to predict the $\mathrm{[MASK]}$ token, akin to the masked language model task.

\subsubsection{Text Reconstruction with MLLM}
However, the textual descriptions in the original corpus contain numerous errors. For example, for the entity ``\textit{Wither}'', the textual description in the corpus is ``\textit{2009 EP by Dream Theater}''. But in analogical reasoning questions, its original meaning is used, namely ``\textit{shrivel or fade}''. 

Therefore, we leverage the image content understanding and text generation capabilities of MLLM to generate descriptive texts for the given entities. For example, the reconstructed textual description of ``\textit{Wither}'' is ``\textit{a spell that causes a target's life force to dwindle.}'', which is close to its original meaning. We input multimodal analogical reasoning problems into MLLM as natural language. Taking the $(I_h,T_t):(I_q,?)$ problem as an example, MLLM needs to find the entities $e_h$ and $e_q$ corresponding to $I_h$ and $I_q$, and construct descriptive texts $T_h$ and $T_q$. We also require MLLM to reconstruct the textual description $T_t$ corresponding to $e_t$. Furthermore, MLLM needs to understand the analogical reasoning problem, describe the relation between $e_h$ and $e_t$, and generate $T_r$. The analogy question and textual descriptions are then incorporated into a unified prompt template. The final step involves feeding the prompt into MPT methods.

\subsection{MLLM as a Predictor}
\begin{figure}[htbp]
\begin{center}
\includegraphics[width=0.8\linewidth]{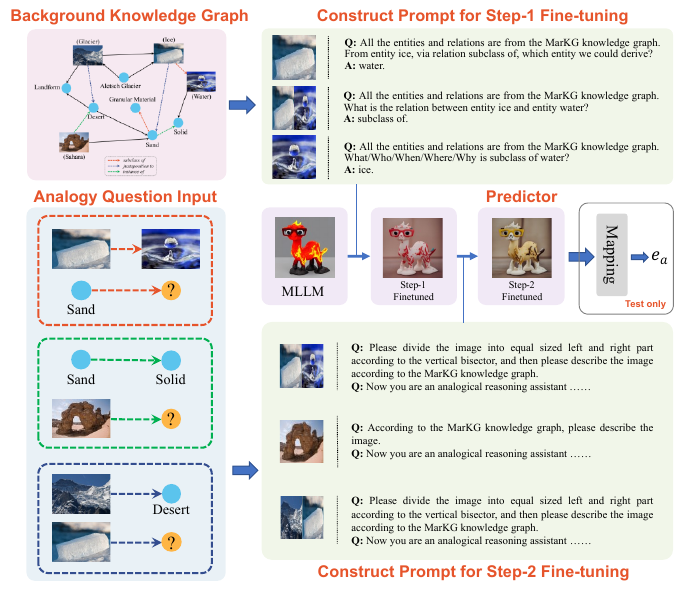} 
\caption{MLLM as a predictor.}
\label{fig:m2}
\end{center}
\end{figure}

We also explore the feasibility of using MLLM to perform multimodal analogical reasoning task and propose a two-step fine-tuning framework. The overall process is illustrated in Figure \ref{fig:m2}.

\subsubsection{Preparation}
Due to structural limitations, existing MLLMs perform poorly when handling multiple image inputs. This hinders MLLMs from directly adapting to the multimodal analogical reasoning task, like $(I_h, T_t):(I_q, ?)$. Therefore, we propose a simple approach to address this issue by combining two input images side by side into a single image. We use $Combine(I_1, I_2)$ to denote this image.

\subsubsection{Step-1 Fine-tuning}
In step-1 fine-tuning, our goal is to enable MLLM to learn the triplet information in the background knowledge graph. Specifically, we construct three tasks for each triplet: head entity prediction, relation prediction, and tail entity prediction. These tasks are then input into the MLLM in natural language as following form:
\begin{align}  
    Prompt_h &= I_h \; | \;  Task_h \\
    Prompt_r &= Combine(I_h, I_t) \; | \;  Task_r \\
    Prompt_t &= I_t \; | \;  Task_t 
\end{align}
where $Task$ represents the textual description of each task. Taking the $(ice, class\; of, water)$ triplet as an example, the prompts for the three types of prediction tasks are given in Figure \ref{fig:m2}.

\subsubsection{Step-2 Fine-tuning}
To guide MLLM in performing multimodal analogical reasoning task, we construct prompts in the form of multi-turn dialogues. For $(I_h, T_t):(I_q, ?)$ problem, the prompt takes the following form:
\begin{align}  
    Prompt_1 &= Combine(I_h, I_q) \; | \;  Question_1 \\
    Prompt_2 &= Question_2
\end{align}
where $Question_1$ is a question about image understanding, $Question_2$ is the multimodal analogical reasoning question.

Specifically, we first require the MLLM to comprehend the given images in $Prompt_1$ and describe the entities corresponding to the image subjects. Next, we construct a natural language description for the multimodal analogical reasoning question. Then, we formulate instructions for predicting implicit relation and answering the analogical reasoning question from the given 10 options. Inevitably, MLLM may generate some entities and relations that do not exist in background knowledge graph due to the \textit{hallucination} problem. Therefore, we introduce a mapping module to calculate the cosine similarity between the output and entities/relations in background knowledge graph. We select the one with the highest similarity score as the final answer. 

\section{Experiments}
In the following section, we use \textit{Explainer} and \textit{Predictor} to represent \textit{MLLM as an explainer} and \textit{MLLM as a predictor}, respectively.

\subsection{Experimental Settings}
\subsubsection{Main Datasets}
We pre-train baselines and perform the step-1 fine-tuning of \textit{Predictor} on MarKG \cite{zhang2023multimodal}. We preform the step-2 fine-tuning of \textit{Predictor} and evaluate all methods on MARS \cite{zhang2023multimodal}. MarKG is a background multimodal knowledge graph with 34,420 triplets, which is collected from Wikidata. It aims to provide prior knowledge of analogous entities and relations. The MARS dataset serves as the training and evaluation resource for the multimodal analogical reasoning task. The dataset contains 10,685 training questions, 1,228 validation questions, and 1,415 test questions.
\subsubsection{MBARD Dataset}
The MARS dataset focuses on analogical reasoning between noun entities, which deviates from real-world scenarios. In reality, we often encounter analogical reasoning between verbs and nouns, such as $(scrub,brush):(stir,spoon)$. Furthermore, analogical reasoning is commonly applied in unfamiliar contexts. Therefore, we construct MBARD, a dataset to evaluate the analogical reasoning capability of MLLM in zero-shot scenarios. The task format of MBARD diverges from MARS in two aspects:
\begin{itemize}
    \item It exclusively involves verb-noun pair analogical reasoning, which is closer to real-life scenarios and the entities have not appeared in training;
    \item The task format is $(T_h,I_t):(T_q,?)$, which has not been seen during training.
\end{itemize}

The analogy examples of MBARD dataset are derived from the BARD dataset \cite{fulda2017harvesting}, selecting a total of 1,000 analogical examples. The images corresponding to the noun entities are obtained through web crawling. 

\subsubsection{Base Model and Baselines}

We use LLaVA \cite{liu2023llava} as our base model to implement \textit{Explainer}. In \textit{Predictor}, we employ LoRA \cite{DBLP:conf/iclr/HuSWALWWC22} for fine-tuning.  We use MiniLM-v2 \cite{reimers-2019-sentence-bert} as the mapping module.

We consider two categories of methods as our baselines, namely MKGE methods and MPT methods. MKGE methods include IKRL \cite{Xie_Liu_Luan_Sun_2017}, TransAE \cite{Wang_Li_Li_Zeng_2019}, and RSME \cite{Wang_Wang_Yang_Zhang_Chen_Qi_2021}. Moreover, we select MPT methods including VisualBERT \cite{li2019visualbert}, ViLT \cite{kim2021vilt}, ViLBERT \cite{lu2019vilbert}, FLAVA \cite{singh2022flava} and MKGformer \cite{chen2022hybrid} as the strong baselines for \textit{Explainer}. We select MLLMs including VisualGLM \cite{du2022glm}, LLaVA \cite{liu2023llava}, MiniCPM-V 2.0 \cite{yu2024rlaifv}, Qwen-VL-Chat \cite{Qwen-VL}, mPLUG-Owl 2 \cite{ye2023mplugowl2} and internVL 2 \cite{chen2023internvl} as baselines for \textit{Predictor}.

\subsubsection{Metrics}
Following the work of  \cite{zhang2023multimodal}, we  use Hits@k (ratio of top k valid entities) and Mean Reciprocal Rank (MRR) as our evaluation metrics. Due to the output constraints of MLLM, when evaluating the \textit{Predictor}, we focus on Hits@1, or accuracy in other words.

\begin{table*}[htbp]
\caption{Main results (\%) for \textit{Explainer} on MARS. Acc refers to accuracy. *The results are derived from  \cite{zhang2023multimodal}. }
\centering  
\renewcommand\arraystretch{1}
\begin{tabular}{clccccc}
\toprule
\multicolumn{2}{c}{\textbf{Method}}  & \textbf{Hits@1/Acc} & \textbf{Hits@3} & \textbf{Hits@5} & \textbf{Hits@10} & \textbf{MRR}   \\ \midrule
\multirow{3}{*}{MKGE} & IKRL$^{*}$ \cite{Xie_Liu_Luan_Sun_2017}            & 25.4          & 28.5          & 29.0          & 30.4           & 27.4          \\
&TransAE$^{*}$ \cite{Wang_Li_Li_Zeng_2019}         & 20.3          & 23.3          & 24.1          & 25.3           & 22.3          \\
&RSME$^{*}$ \cite{Wang_Wang_Yang_Zhang_Chen_Qi_2021}           & 25.5          & 27.4          & 28.2          & 29.1           & 26.8          \\ \midrule
\multirow{5}{*}{MPT}&VisualBERT$^{*}$ \cite{li2019visualbert}      & 26.1          & 29.2          & 30.8          & 32.1           & 28.4          \\
&ViLT$^{*}$ \cite{kim2021vilt}            & 24.5          & 27.5          & 28.7          & 30.3           & 26.6          \\
&ViLBERT$^{*}$ \cite{lu2019vilbert}         & 25.6          & 31.2          & 32.7          & 34.7           & 29.2          \\
&FLAVA$^{*}$ \cite{singh2022flava}           & 26.4          & 30.3          & 30.9          & 31.9           & 28.8          \\
&MKGformer$^{*}$ \cite{chen2022hybrid}       & 30.1          & 36.7          & 38.0          & 40.8           & 34.1          \\ \midrule
\multirow{2}{*}{\textit{Explainer}}& + FLAVA              & $\mathbf{33.3}_{\uparrow 6.9}$ & $\mathbf{38.3}_{\uparrow 8.0}$ & 
$39.9_{\uparrow 9.0}$          & 
$41.4_{\uparrow 9.5}$           & 
$36.4_{\uparrow 7.6}$          \\
& + MKGformer               & $32.4_{\uparrow 2.3}$          & $\mathbf{38.3}_{\uparrow 1.6}$ & $\mathbf{40.3}_{\uparrow 2.3}$ & $\mathbf{43.4}_{\uparrow 2.6}$  & $\mathbf{36.6}_{\uparrow 2.5}$ \\ 

\bottomrule
\end{tabular}
\label{tab:result}
\end{table*}

\subsection{Main Results}

We compare our methods with baselines on MARS and report the results in Table \ref{tab:result} and \ref{tab:predictor}. From the results, we observe that our methods can achieve state-of-the-art performance compared with baselines. 
Specifically, compared with the strong baseline MKGformer, \textit{Explainer}+MKGformer performs better, exceeding MKGformer by 1.6\%-2.6\% in all five metrics. Furthermore, \textit{Explainer}+FLAVA achieves a remarkable improvement of 6.9\%-9.5\% across all metrics compared to FLAVA.
For MLLMs, analogical reasoning is a fundamental emergent ability, so most MLLMs can exhibit high accuracy without instruction fine-tuning. Our \textit{Predictor} framework outperforms all MLLM baselines. Among them, \textit{Predictor}(LLaVA) achieves an accuracy of 56.2\%, which significantly exceeds other models. Based on these observations, two main findings can be summarized:
\textbf{\textit{Finding 1}}: \textit{Explainer} and \textit{Predictor} can effectively enhance the multimodal analogical reasoning ability of existing models.
\textit{\textbf{Finding 2}}: MLLM can understand and solve the multimodal analogical reasoning task.

\begin{table}[htbp]
\caption{Main results (\%) for \textit{Predictor} on MARS. }
\centering
\begin{tabular}{lll}
\toprule
\textbf{Method}            & \textbf{\# Param} & \textbf{Accuracy(\%)} \\ \midrule
VisualGLM         & 6B         & 6.24         \\
LLaVA         & 7B         & 43.39        \\
MiniCPM-V 2.0     & 2.8B       & 37.29        \\
Qwen-VL-Chat      & 7B         & 36.12        \\
mPLUG-Owl 2       & 7B         & 36.12        \\ 
internVL 2  & 8B & 45.69 \\
\midrule
\textit{Predictor}(VisualGLM) & 6B         & $\mathbf{35.61}_{\uparrow 29.37}$        \\
\textit{Predictor}(LLaVA) & 7B         & $\mathbf{56.20}_{\uparrow 12.81}$      \\ \bottomrule

\end{tabular}
\label{tab:predictor}
\end{table}

\subsection{Results of Implicit Relation Inference}
In multimodal analogical reasoning task, the relation between $e_h$ and $e_t$ is not explicitly provided. Therefore, we want to investigate whether the proposed methods can accurately predict the implicit relation. As shown in Table \ref{tab:relation}, existing methods perform poorly in predicting implicit relation, and even our \textit{Explainer} method does not achieve state-of-the-art results in all metrics. 
However, as shown in Table \ref{tab:pre_relation}, MLLMs have a higher accuracy in predicting implicit relation. In particular, our \textit{Predictor}(VisualGLM) achieves an accuracy of 47.2\%, significantly higher than other methods.
The experimental results indicate that MKGE models can hardly predict implicit relation accurately, while MLLMs can provide more accurate reasoning paths. We believe that this phenomenon is attributed to the emergent analogical reasoning capability of MLLMs.

\begin{table}[htbp]
  \centering
  \begin{minipage}{0.5\textwidth}
    
\caption{Implicit relation inference results (\%) on MARS for \textit{Explainer}.}
\label{tab:relation}
    \centering
    \resizebox{\linewidth}{!}{
\begin{tabular}{llll}
\toprule
\textbf{Method}     & \textbf{Hits@1/Acc} & \textbf{Hits@3} & \textbf{Hits@5} \\ \midrule
IKRL       &  6.6        & 16.0  & 23.4  \\
VisualBERT &  3.8         & 10.7  & 18.1  \\
FLAVA      &  1.9      & 7.8  & 58.7  \\
MKGformer  &  0.5        & 4.9  & 20.9  \\ \midrule
\textit{Explainer}+FLAVA          & 2.1    & 13.9     &  55.0    \\
\textit{Explainer}+MKGformer          & 1.6    & 4.3      & 24.8      \\ 
\bottomrule
\end{tabular}}
  \end{minipage}%
  \hspace{0.05\textwidth}
  \begin{minipage}{0.4\textwidth}
    \caption{Implicit relation inference results (\%) on MARS for \textit{Predictor}.}
\label{tab:pre_relation}
    \centering
\resizebox{\linewidth}{!}{
\begin{tabular}{lll}
\toprule
\textbf{Method}            & \textbf{\# Param} & \textbf{Accuracy(\%)} \\ \midrule
VisualGLM         & 6B         & 5.80         \\
LLaVA             & 7B         & 25.55        \\
MiniCPM-V 2.0     & 2.8B       & 21.81        \\
Qwen-VL-Chat      & 7B         & 36.12        \\
mPLUG-Owl 2       & 7B         & 8.14         \\ 
internVL 2        & 8B         & 28.11         \\\midrule
\textit{Predictor}(VisualGLM) & 6B         & \textbf{47.20}        \\
\textit{Predictor}(LLaVA) & 7B         & \textbf{44.12}        \\ \bottomrule
\end{tabular}}
  \end{minipage}
\end{table}

\subsection{Q\&A, Multiple-choice or True/False?}

\begin{figure}[htbp]
    \begin{minipage}{0.55\textwidth}
        \centering
        \includegraphics[width=\linewidth]{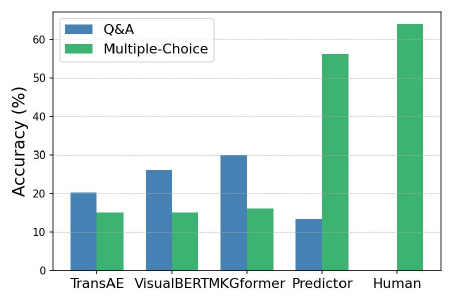}
        \caption{Q\&A and multiple-choice evaluation. * denotes \textit{Predictor}.}
        \label{fig:mc}
    \end{minipage}%
    \hfill
    \begin{minipage}{0.45\textwidth}
        \centering
        \includegraphics[width=\linewidth]{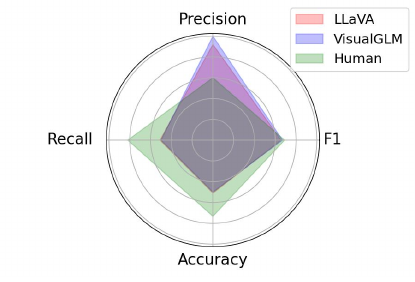}
        \caption{True/False evaluation.}
        \label{fig:judge}
    \end{minipage}
\end{figure}

We further investigate the impact of different prompt formats on \textit{Predictor}. We primarily focus on three prompt modes: Q\&A, multiple-choice, and True/False. To further highlight the task's difficulty, we introduce human evaluation in the multiple-choice mode. The Q\&A mode is the default mode for the baseline models and does not provide answer options. Taking the LLaVA-based model as an example, the experimental results for the Q\&A and multiple-choice modes are shown in Figure \ref{fig:mc}. It is evident that the baseline methods perform less favorably in the multiple-choicee mode compared to the Q\&A mode, while the \textit{Predictor}'s performance in the multiple-choice mode surpasses other methods, even approaching human-level performance. The True/False mode aims to determine whether the given analogical reasoning example is valid, with experimental results shown in Table \ref{fig:judge}. We find that \textit{Predictor} tends to provide a ``True'' response, exhibiting poor performance. We believe that this is due to \textit{Sycophancy} behavior of MLLM \cite{wei2023simple}. In conclusion, based on the current work, the multiple-choice mode appears to be the most effective mode for \textit{Predictor}. 

\subsection{Zero-shot Evaluation on MBARD}
\begin{figure}[htbp]
\begin{center}
\includegraphics[width=0.85\linewidth]{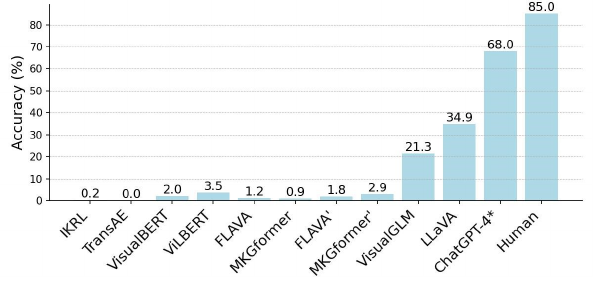} 
\caption{Zero-shot evaluation. * denotes \textit{Explainer}. Pre($\cdot$) denotes \textit{Predictor}. }
\label{fig:zero}
\end{center}
\end{figure}

In real-world scenarios, humans often rely on analogical reasoning to gain initial insights in unfamiliar situations. Therefore, we conduct experiments for zero-shot multimodal analogical reasoning on our MBARD dataset. We also introduce ChatGPT-4 and human evaluations for comparison. The experimental results are illustrated in Figure \ref{fig:zero}. MLLMs  demonstrate a certain level of zero-shot multimodal analogical reasoning capability, with ChatGPT-4 exhibiting the best performance, achieving an accuracy of 68.0\%. \textit{Predictor}(LLaVA) exhibits accuracy close to that of ChatGPT-4, significantly outperforming other methods. In contrast, baseline models like MKGformer are unable to complete this challenging task. Hence, the experimental results suggest that MLLMs can preliminarily perform multimodal analogical reasoning in zero-shot scenarios.

\begin{table}[htbp]
\caption{Ablation studies.}
\centering
\begin{tabular}{lcccc}
\toprule
\textbf{Method}          & \textbf{Hits@1/Acc} & \textbf{Hits@5} & \textbf{Hits@10} & \textbf{MRR} \\ \midrule
\textit{Explainer}+MKGformer& 32.4           & \textbf{40.3}           & \textbf{43.4}            & \textbf{36.6}        \\
\; w/o $T_r$   & 30.3           & 37.9           & 41.7            & 34.4        \\
\; w/o $T_e$   & 28.9               & 37.7               & 39.9                & 33.1            \\
\; w/o $T_r+T_e$ & 28.6           & 35.0           & 37.2            & 31.9        \\ \midrule

\textit{Predictor(LLaVA)}                        & \textbf{56.2}           & -               & -                & -                 \\
\; w/o $\mathcal{M}$   & 54.4 &-&-&-    \\ 
\; w/o ft1   & 24.3               & -               & -                & -            \\ 
\; w/o ft2   & 23.2               & -               & -                & -            \\ 
\; w/o ft1+ft2   & 1.1               & -               & -                & -            \\ 
\bottomrule
\end{tabular}
\label{tab:ablation}
\end{table}

\subsection{Ablation Studies}
In this section, we compare our complete framework with several variants: ``w/o $T_r$'' is \textit{Explainer} without textual description of relation; ``w/o $T_e$'' is  \textit{Explainer} without textual descriptions of entities $e_h$, $e_t$ and $e_q$; ``w/o $\mathcal{M}$'' is \textit{Predictor} without mapping module; ``w/o ft1'' is \textit{Predictor} without step-1 fine-tuning; ``w/o ft2'' is \textit{Predictor} without step-2 fine-tuning. Taking \textit{Explainer} + MKGformer and \textit{Predictor}(LLaVA) as examples, the results are shown in Table \ref{tab:ablation}. We observe that discarding any component results in worse performance for both  \textit{Explainer} and \textit{Predictor}. Specifically, for \textit{Explainer}, using descriptive texts generated by \textit{Explainer} for entities and relations can effectively enhance model performance. Notably, the contribution of $T_e$ is more significant. It demonstrates the effectiveness of our proposed unified prompt template and \textit{Explainer} method. For \textit{Predictor}, the results strongly highlight the importance of the two-step fine-tuning framework.

\begin{figure*}[htbp]
\begin{center}
\includegraphics[width=0.9\linewidth]{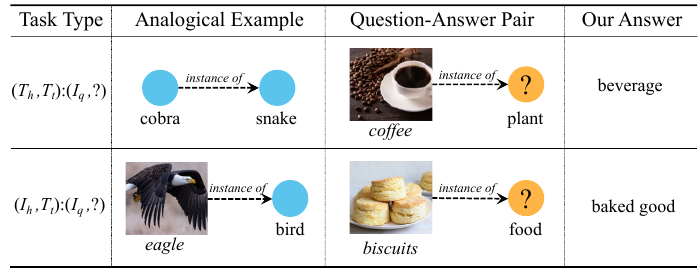} 
\caption{Case examples of MARS.}
\label{fig:case}
\end{center}
\end{figure*}

\subsection{Analysis of Errors}
To gain a deeper understanding of MLLM's multimodal analogical reasoning ability, we conduct an analysis of several error examples. As shown in Figure \ref{fig:case}, for the analogical reasoning problem $(cobra, snake):(coffee, ?)$, both the correct answer $plant$ and the \textit{Predictor}'s answer $beverage$ are reasonable from a human perspective. Similarly, for the analogical reasoning problem $(eagle, bird):(biscuits, ?)$, the answer $baked \; good$ given by the \textit{Predictor} is also valid. In addition, we also find some perplexing problems in the MARS dataset, such as $(fish,school):(quality\; control,?)$. It is incomprehensible to humans and to the \textit{Predictor}. Observations in Figure \ref{fig:case} indicate that the practical performance of the \textit{Predictor} may be superior to what are presented in the main results. To provide an accurate model performance, we plan to optimize the dataset in the future.

\section{Conclusion}
In this paper, we explore the multimodal analogical reasoning capability of the Multimodal Large Language Model (MLLM). We propose two advanced frameworks: \textit{MLLM as an explainer} and \textit{MLLM as a predictor}. \textit{MLLM as an explainer} focuses on template reconstruction and analogical reasoning problem comprehension, aiming to enhance the analogical reasoning abilities of existing methods. \textit{MLLM as a predictor}, on the other hand, investigates the analogical reasoning capabilities of MLLM itself. Our experiments demonstrate that both frameworks achieve state-of-the-art results, providing initial evidence that MLLM can perform multimodal analogical reasoning task effectively. We believe that our work has the potential to inspire research on the cognitive abilities of MLLMs. Moreover, in future work, we intend to delve deeper into understanding the specific types of analogical reasoning problems that MLLMs are particularly adept at addressing.

\subsubsection{\ackname} This research is supported by the National Key R\&D Program of China (No. 2023YFC3303800).

\bibliographystyle{splncs04}
\bibliography{6250}

\begin{thebibliography}{10}
\providecommand{\url}[1]{\texttt{#1}}
\providecommand{\urlprefix}{URL }
\providecommand{\doi}[1]{https://doi.org/#1}

\bibitem{Qwen-VL}
Bai, J., Bai, S., Yang, S., Wang, S., Tan, S., Wang, P., Lin, J., Zhou, C., Zhou, J.: Qwen-vl: A versatile vision-language model for understanding, localization, text reading, and beyond. arXiv preprint arXiv:2308.12966  (2023)

\bibitem{bordes2013translating}
Bordes, A., Usunier, N., Garcia-Duran, A., Weston, J., Yakhnenko, O.: Translating embeddings for modeling multi-relational data. Advances in neural information processing systems  \textbf{26} (2013)

\bibitem{brown2020language}
Brown, T., Mann, B., Ryder, N., Subbiah, M., Kaplan, J.D., Dhariwal, P., Neelakantan, A., Shyam, P., Sastry, G., Askell, A., et~al.: Language models are few-shot learners. Advances in neural information processing systems  \textbf{33},  1877--1901 (2020)

\bibitem{cattell1971abilities}
Cattell, R.B.: Abilities: Their structure, growth, and action.  (1971)

\bibitem{chen2022hybrid}
Chen, X., Zhang, N., Li, L., Deng, S., Tan, C., Xu, C., Huang, F., Si, L., Chen, H.: Hybrid transformer with multi-level fusion for multimodal knowledge graph completion. In: Proceedings of the 45th International ACM SIGIR Conference on Research and Development in Information Retrieval. pp. 904--915 (2022)

\bibitem{chen2023internvl}
Chen, Z., Wu, J., Wang, W., Su, W., Chen, G., Xing, S., Zhong, M., Zhang, Q., Zhu, X., Lu, L., Li, B., Luo, P., Lu, T., Qiao, Y., Dai, J.: Internvl: Scaling up vision foundation models and aligning for generic visual-linguistic tasks. arXiv preprint arXiv:2312.14238  (2023)

\bibitem{du2022glm}
Du, Z., Qian, Y., Liu, X., Ding, M., Qiu, J., Yang, Z., Tang, J.: Glm: General language model pretraining with autoregressive blank infilling. In: Proceedings of the 60th Annual Meeting of the Association for Computational Linguistics (Volume 1: Long Papers). pp. 320--335 (2022)

\bibitem{dunbar2012scientific}
Dunbar, K.N., Klahr, D.: Scientific thinking and reasoning  (2012)

\bibitem{ethayarajh2018towards}
Ethayarajh, K., Duvenaud, D., Hirst, G.: Towards understanding linear word analogies. arXiv preprint arXiv:1810.04882  (2018)

\bibitem{fulda2017harvesting}
Fulda, N., Tibbetts, N., Brown, Z., Wingate, D.: Harvesting common-sense navigational knowledge for robotics from uncurated text corpora. In: Conference on Robot Learning. pp. 525--534. PMLR (2017)

\bibitem{gentner2012analogical}
Gentner, D., Smith, L., Ramachandran, V.: Analogical reasoning. encyclopedia of human behavior (2012)

\bibitem{glynn1989analogical}
Glynn, S.M., Britton, B.K., Semrud-Clikeman, M., Muth, K.D.: Analogical reasoning and problem solving in science textbooks. Handbook of creativity pp. 383--398 (1989)

\bibitem{DBLP:conf/iclr/HuSWALWWC22}
Hu, E.J., Shen, Y., Wallis, P., Allen{-}Zhu, Z., Li, Y., Wang, S., Wang, L., Chen, W.: Lora: Low-rank adaptation of large language models. In: The Tenth International Conference on Learning Representations, {ICLR} 2022, Virtual Event, April 25-29, 2022. OpenReview.net (2022)

\bibitem{hu2021stratified}
Hu, S., Ma, Y., Liu, X., Wei, Y., Bai, S.: Stratified rule-aware network for abstract visual reasoning. In: Proceedings of the AAAI Conference on Artificial Intelligence. vol.~35, pp. 1567--1574 (2021)

\bibitem{hu2023thought}
Hu, S., Clune, J.: Thought cloning: Learning to think while acting by imitating human thinking. arXiv preprint arXiv:2306.00323  (2023)

\bibitem{kim2021vilt}
Kim, W., Son, B., Kim, I.: Vilt: Vision-and-language transformer without convolution or region supervision. In: International Conference on Machine Learning. pp. 5583--5594. PMLR (2021)

\bibitem{li2019visualbert}
Li, L.H., Yatskar, M., Yin, D., Hsieh, C.J., Chang, K.W.: Visualbert: A simple and performant baseline for vision and language. arXiv preprint arXiv:1908.03557  (2019)

\bibitem{liu2023llava}
Liu, H., Li, C., Wu, Q., Lee, Y.J.: Visual instruction tuning. arXiv preprint arXiv:2304.08485  (2023)

\bibitem{lu2019vilbert}
Lu, J., Batra, D., Parikh, D., Lee, S.: Vilbert: Pretraining task-agnostic visiolinguistic representations for vision-and-language tasks. Advances in neural information processing systems  \textbf{32} (2019)

\bibitem{malkinski2022deep}
Ma{\l}ki{\'n}ski, M., Ma{\'n}dziuk, J.: Deep learning methods for abstract visual reasoning: A survey on raven's progressive matrices. arXiv preprint arXiv:2201.12382  (2022)

\bibitem{meguro2020effects}
Meguro, Y.: The effects of individual differences in field dependence/independence and analogical reasoning for l2 instruction. System  \textbf{94},  102296 (2020)

\bibitem{mitchell2021abstraction}
Mitchell, M.: Abstraction and analogy-making in artificial intelligence. Annals of the New York Academy of Sciences  \textbf{1505}(1),  79--101 (2021)

\bibitem{novick2005problem}
Novick, L.R., Bassok, M.: Problem Solving. Cambridge University Press (2005)

\bibitem{reimers-2019-sentence-bert}
Reimers, N., Gurevych, I.: Sentence-bert: Sentence embeddings using siamese bert-networks. In: Proceedings of the 2019 Conference on Empirical Methods in Natural Language Processing. Association for Computational Linguistics (11 2019)

\bibitem{singh2022flava}
Singh, A., Hu, R., Goswami, V., Couairon, G., Galuba, W., Rohrbach, M., Kiela, D.: Flava: A foundational language and vision alignment model. In: Proceedings of the IEEE/CVF Conference on Computer Vision and Pattern Recognition. pp. 15638--15650 (2022)

\bibitem{touvron2023llama}
Touvron, H., Lavril, T., Izacard, G., Martinet, X., Lachaux, M.A., Lacroix, T., Rozi{\`e}re, B., Goyal, N., Hambro, E., Azhar, F., et~al.: Llama: Open and efficient foundation language models. arXiv preprint arXiv:2302.13971  (2023)

\bibitem{ushio2021bert}
Ushio, A., Espinosa-Anke, L., Schockaert, S., Camacho-Collados, J.: Bert is to nlp what alexnet is to cv: can pre-trained language models identify analogies? arXiv preprint arXiv:2105.04949  (2021)

\bibitem{Wang_Wang_Yang_Zhang_Chen_Qi_2021}
Wang, M., Wang, S., Yang, H., Zhang, Z., Chen, X., Qi, G.: Is visual context really helpful for knowledge graph? a representation learning perspective. In: Proceedings of the 29th ACM International Conference on Multimedia (Oct 2021)

\bibitem{Wang_Li_Li_Zeng_2019}
Wang, Z., Li, L., Li, Q., Zeng, D.: Multimodal data enhanced representation learning for knowledge graphs. In: 2019 International Joint Conference on Neural Networks (IJCNN) (Jul 2019)

\bibitem{webb2023emergent}
Webb, T., Holyoak, K.J., Lu, H.: Emergent analogical reasoning in large language models. Nature Human Behaviour pp. 1--16 (2023)

\bibitem{wei2023simple}
Wei, J., Huang, D., Lu, Y., Zhou, D., Le, Q.V.: Simple synthetic data reduces sycophancy in large language models. arXiv preprint arXiv:2308.03958  (2023)

\bibitem{Xie_Liu_Luan_Sun_2017}
Xie, R., Liu, Z., Luan, H., Sun, M.: Image-embodied knowledge representation learning. In: Proceedings of the Twenty-Sixth International Joint Conference on Artificial Intelligence (Aug 2017)

\bibitem{yasunaga2023large}
Yasunaga, M., Chen, X., Li, Y., Pasupat, P., Leskovec, J., Liang, P., Chi, E.H., Zhou, D.: Large language models as analogical reasoners. arXiv preprint arXiv:2310.01714  (2023)

\bibitem{ye2023mplugowl2}
Ye, Q., Xu, H., Ye, J., Yan, M., Hu, A., Liu, H., Qian, Q., Zhang, J., Huang, F., Zhou, J.: mplug-owl2: Revolutionizing multi-modal large language model with modality collaboration (2023)

\bibitem{yin2023survey}
Yin, S., Fu, C., Zhao, S., Li, K., Sun, X., Xu, T., Chen, E.: A survey on multimodal large language models. arXiv preprint arXiv:2306.13549  (2023)

\bibitem{yu2024rlaifv}
Yu, T., Zhang, H., Yao, Y., Dang, Y., Chen, D., Lu, X., Cui, G., He, T., Liu, Z., Chua, T.S., Sun, M.: Rlaif-v: Aligning mllms through open-source ai feedback for super gpt-4v trustworthiness. arXiv preprint arXiv:2405.17220  (2024)

\bibitem{zeng2022glm}
Zeng, A., Liu, X., Du, Z., Wang, Z., Lai, H., Ding, M., Yang, Z., Xu, Y., Zheng, W., Xia, X., et~al.: Glm-130b: An open bilingual pre-trained model. arXiv preprint arXiv:2210.02414  (2022)

\bibitem{zhang2019raven}
Zhang, C., Gao, F., Jia, B., Zhu, Y., Zhu, S.C.: Raven: A dataset for relational and analogical visual reasoning. In: Proceedings of the IEEE/CVF conference on computer vision and pattern recognition. pp. 5317--5327 (2019)

\bibitem{zhang2023multimodal}
Zhang, N., Li, L., Chen, X., Liang, X., Deng, S., Chen, H.: Multimodal analogical reasoning over knowledge graphs. In: The Eleventh International Conference on Learning Representations (2023)

\bibitem{zhao2022mose}
Zhao, Y., Cai, X., Wu, Y., Zhang, H., Zhang, Y., Zhao, G., Jiang, N.: Mose: Modality split and ensemble for multimodal knowledge graph completion. arXiv preprint arXiv:2210.08821  (2022)

\end{thebibliography}

\end{document}